# Body Composition Estimation Based on Multimodal Multi-task Deep Neural Network

Subas Chhatkuli, Iris Jiang, and Kyohei Kamiyama
*Bodygram Japan K.K.*

**Abstract**

In addition to body weight and Body Mass Index (BMI), body composition is an essential data point that allows people to understand their overall health and body fitness. However, body composition is largely made up of muscle, fat, bones, and water, which makes estimation not as easy and straightforward as measuring body weight. In this paper, we introduce a multimodal multi-task deep neural network to estimate body fat percentage and skeletal muscle mass by analyzing facial images in addition to a person's height, gender, age, and weight information. Using a dataset representative of demographics in Japan, we confirmed that the proposed approach performed better compared to the existing methods. Moreover, the multi-task approach implemented in this study is also able to grasp the negative correlation between body fat percentage and skeletal muscle mass gain/loss.

## 1. INTRODUCTION

Being conscious about one's body condition is crucial in maintaining a healthy lifestyle. One of the very simple and easy ways to understand overall body condition could be by regularly monitoring body weight and/or BMI. However, body weight and BMI alone do not tell everything about overall health and fitness because they do not account for the difference in density between fat and muscle. People with more muscle mass will often land in the "overweight" or "obese" BMI category based on height and weight, even though they may be healthy. Earlier research suggests that in addition to body weight and BMI, body composition such as body fat percentage (PBF) and skeletal muscle mass (SMM) are very important factors that determine overall good health and body fitness. The study [1] found that diabetes patients with a high PBF in addition to low BMI might develop sarcopenia, which is a condition that causes decrease in muscle mass and decline in physical ability, often negatively influencing quality of life. [2] Evaluated the relationship between components of body composition and mortality in patients with cardiovascular disease. They found the specific subgroup of high muscle and low fat mass had the lowest mortality rate compared to other body composition sub types. [3] demonstrated that subtotal (total minus head) fat percentage is an important feature in predicting the maximum exercise capacity for adults. In both females as well as males, body fat and age were the most important features in predicting maximum exercise capacity in people over the age of 40 with diabetes. Moreover, studies have shown that increased muscle mass increases skeletal muscle glucose uptake and improves insulin sensitivity [4].

There has also been research to systematically examine the relationship between body size, body shape, age, and SMM. [5] found that body weight, height, waist circumference, and age alone and in combination were significantly correlated with SMM (all, $p < 0.001$).

Apart from body shapes and sizes, BMI and PBF are shown in our faces too. Individuals with lower body fat proportion have a more angular face with relatively narrower cheeks and a pointed chin [6]. Authors [7] performed curve-fitting along lower facial geometry and examined the correlation between fitting parameters and body composition. They found that the fitting parameters were highly correlated with BMI and PBF ($R2 > 0.72$).

Even though PBF and SMM are very important indicators of one's health, measuring them is not as easy as measuring body weight or BMI. Dual Energy X-ray Absorptiometry (DEXA), is very accurate and precise to measure body fat, however it is not cheap and not possible to use at home. Likewise, body composition scales like Inbody, Tanita, and Omron can be used to measure PBF and SMM but these devices are not cheap to buy for home use. Many home-use digital scales offer body fat measurements but are largely inaccurate with one study showing absolute errors of 2.2 - 4.4kg of fat mass [10].

With that in mind, we have developed a method to measure PBF and SMM that can be deployed on a smartphone. Leveraging the power of deep neural networks, we devised a very efficient and highly accurate way of measuring PBF and SMM by analyzing facial images and a person's height, gender, age, weight information. The measurement can be done in a natural background environment irrespective of whether the person is indoor or outdoor.

## 2. MATERIALS AND METHOD

### 2.1 Data acquisition

We collected data in Japan of over 1,500 people of different ages with varying body weights and heights. For the body composition data collection, we used an InBody 270 Body Composition Analyzer (https://inbodyusa.com/). In addition to body composition we also recorded height, gender, age, and body weight of each individual that we scanned. During data acquisition, we also took a fully clothed front and side picture of each subject.

Our data consists of predominantly Japanese people (97%), and demographic bias is reflected in the distributions seen in Figure 1. It is important to account for biases in the dataset as conclusions may not necessarily be generalizable. Our dataset has an almost equal distribution between male and female, 51% and 49% respectively. Figure 1a shows that age ranges from a minimum of 7 years to a maximum of 84 years. The mean and standard deviation of age distribution in our data set is 40.4 years and 12.9 year respectively. Figure 1b shows the distribution of heights, with a mean of 165.4cm and standard deviation of 9.7cm. Figure 1c shows that body weight ranged from a minimum of 18.7kg to a maximum of 141.8kg with mean weight of 63.25kg and standard deviation of 15.9kg.Figure 1d and 1e shows the distribution of SMM and PBF in our data set. The mean values of SMM and PBF are 25.8kg and 25.2% respectively. Likewise, the standard deviation of SMM and PBF were 6.5kg and 8.2%.

Note that normal distribution graphs in this analysis will occasionally have two peaks, which represent the offset between men and women in the dataset.

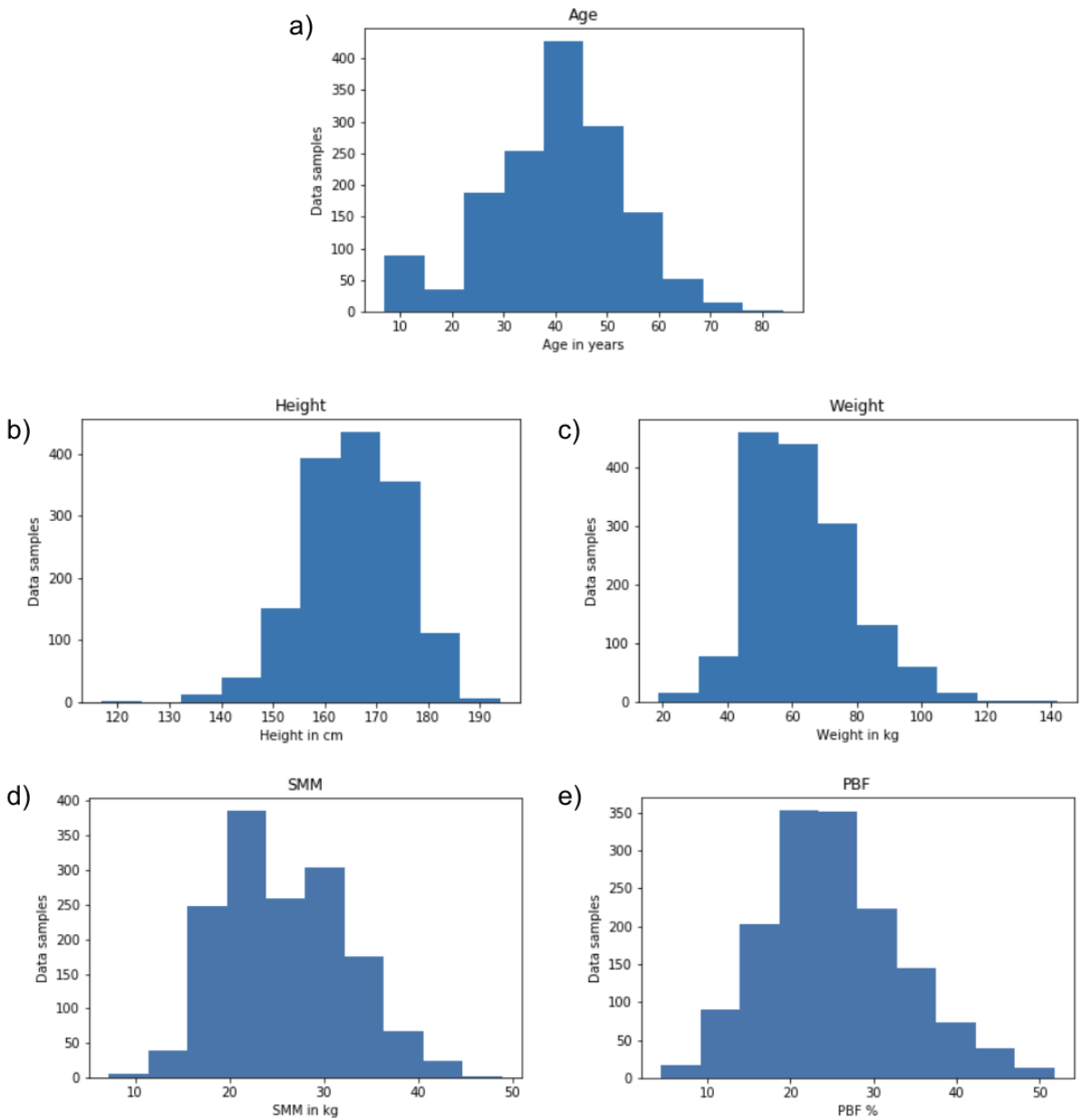

Figure 1: Dataset distributions for (a) Age, (b) Height, (c) Weight, (d) SMM, and (e) PBF

## 2.2 Statistical analysis

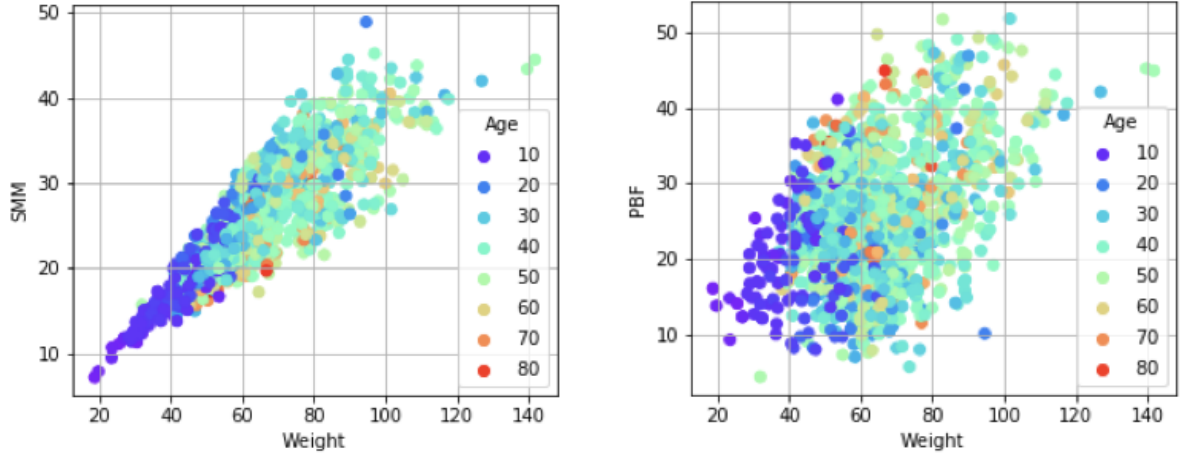

Figure 2: Plot of SMM Vs. body weight (left) and plot of PBF Vs. body weight (right). Here the color in the scatter plot represents age group (i.e. "30" represents people ages 30-39).

On top of the deep neural network based analysis, we also performed a simple statistical analysis. Figure 2 (left) shows a scatter plot of SMM vs body weight and Figure 2 (right) shows a scatter plot of PBF vs body weight. Here, the scatter plots were colored by the age group. From the figure we can confirm a strong correlation between body weight and SMM. Moreover, we can see an interesting trend of more deviation in SMM as the body weight increases (Figure 2, left), suggesting the weaking linear relationship between SMM and body weight as body weight increases. The scatter plot (Figure 2, right) shows some overall trend between body weight and PBF, however the linear relationship between body weight and PBF is not as strong as that between the body weight and SMM.

To objectively quantify the relationship between known body measurement data with SMM and PBF, we performed Pearson's correlation analysis (Figure 3). As seen in Figure 3, The maximum linear correlation of SMM is with body weight (Pearson's correlation = 0.86) and then with height (Pearson's correlation = 0.84). As discussed above, linear correlation between body weight and PBF is lower compared to body weight and SMM. Pearson's correlation coefficient between body weight and PBF was just 0.39.

Moreover, we found that PBF are slightly negatively correlated with SMM (Pearson's correlation coefficient = -0.1). Moreover, males are associated with higher SMM (Pearson's correlation coefficient = 0.7) whereas females are associated with higher PBF (Pearson's correlation coefficient = 0.36).

|      | Race   | Height | Gender | Age   | Weight | SMM   | PBF   |
|------|--------|--------|--------|-------|--------|-------|-------|
| Race | 1      | -0.1   | -0.047 | 0.041 | -0.06  | -0.087| 0.032 |
| Height | -0.1 | 1      | 0.55   | 0.2   | 0.65   | 0.84  | -0.23 |
| Gender | -0.047 | 0.55 | 1      | -0.03 | 0.47   | 0.7   | -0.36 |
| Age  | 0.041  | 0.2    | -0.03  | 1     | 0.32   | 0.23  | 0.25  |
| Weight | -0.06 | 0.65  | 0.47   | 0.32  | 1      | 0.86  | 0.39  |
| SMM  | -0.087 | 0.84   | 0.7    | 0.23  | 0.86   | 1     | -0.11 |
| PBF  | 0.032  | -0.23  | -0.36  | 0.25  | 0.39   | -0.11 | 1     |

Figure 3: Pearson's correlation coefficient

## 2.3 Data preparation and preprocessing

In our multimodal neural network architecture there are six different data input branches. Five branches take images as an input and the sixth branch takes structured data as an input.

The image branches take a full frontal face image and four equal quarters, i.e. upper left (UL), upper right (UR), lower left (LL) and lower right (LR) parts, of the frontal face image. To detect a face from a full frontal image, we utilized an open source library called 'Dlib', which is used to draw a bounding box surrounding a face. On the detected frontal face region of interest, we added a 30% additional margin to include the whole face and a portion of the neck below the chin. The detected face was then cropped from the full frontal image and resized to 128 pixel x 128 pixel. From the full frontal face, four quarters of images (UL, UR, LL, LR) were cropped and then each resized to 128 pixel x 128 pixel. All the images were converted from RGB to grayscale before passing them through the deep neural network. Moreover, during the training phase, slight rotation and gaussian noise were added on the face images for data augmentation purposes.

The next input branch is the structured data branch. In this branch we pass the height, gender, age, and weight information of the person. We also performed a polynomial curve fitting along the person's chin (on the front full photo) and used the fitting parameters as an additional input feature on this branch. During the training phase, we added a gaussian noise on the height and weight category for data augmentation purposes.

The data was randomly split into training (90%) and validation (10%) sets.

## 2.4 Deep neural network architecture

Different parts of a face and other body features like height, gender, age, and weight contain different clues about body composition. However, the influence of these features in determining body composition might not be the same. Hence, we design a multi modal, multi-task deep neural network architecture in which a whole face and each quarter of the whole face and other features like height, gender, age, weight and polynomial fitting parameters are fed through independent parallel networks. These parallel networks learn the parameters for each of these inputs separately. In this way the network learns to weigh the importance of different input features to estimate a body composition. We then concatenate their outputs and pass them through a fully connected shared network before splitting the network to task specific layers. Generally, PBF and SMM are inversely correlated. Hence, other than learning the non linear relationship among these input features to predict a body composition, the shared fully connected layer will help the network to learn this inverse relationship between PBF and SMM too. Moreover, the shared layer also reduces the risk of over fitting [8].

Finally the network is branched into 2 different fully connected task specific layers that learns the parameters to predict body composition namely PBF and SMM as a regression task.

The architecture of the multi input multi task network is shown below (Figure 4).

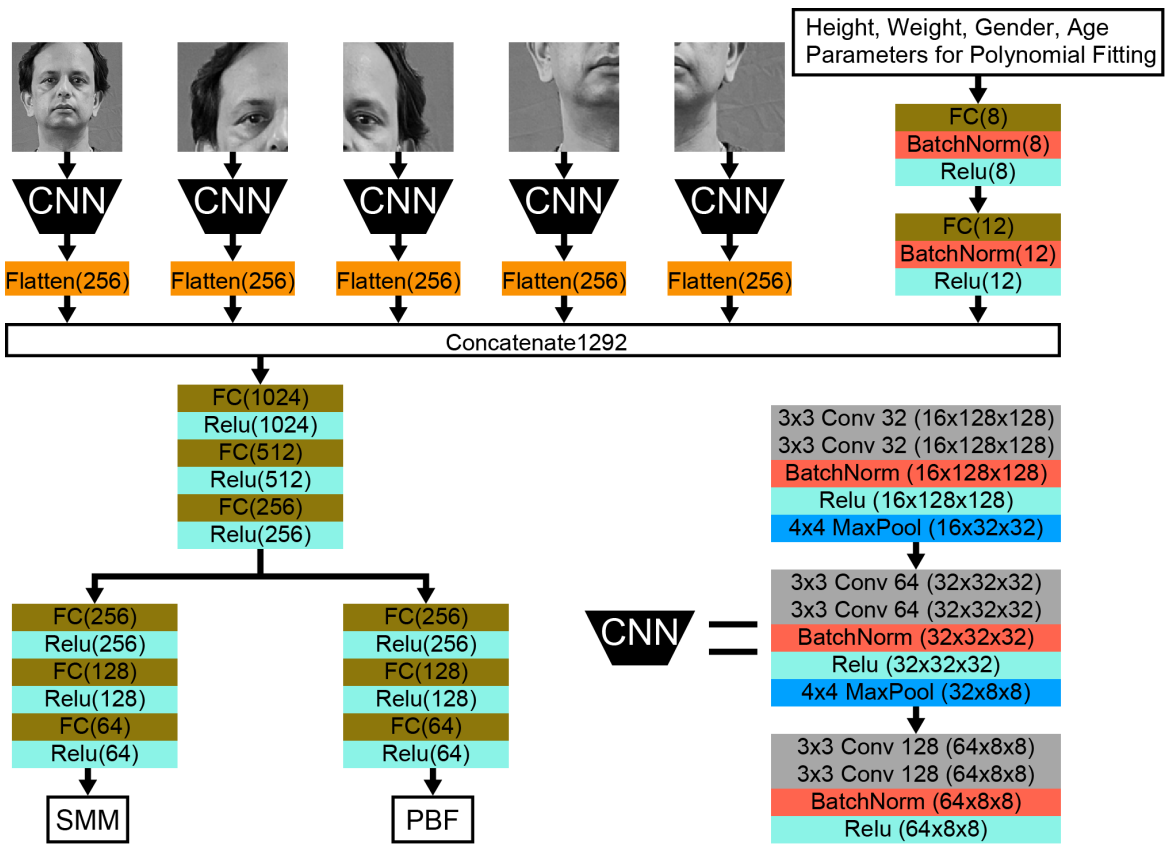

Figure 4: Schematic of AI architecture for body composition estimation

## 3. RESULTS AND DISCUSSION

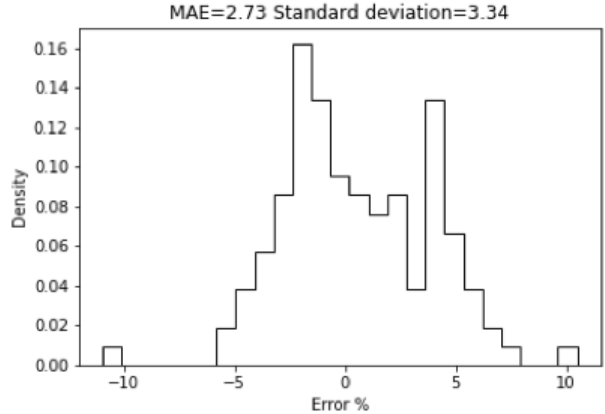

Figure 5: Density plot of PBF estimation error

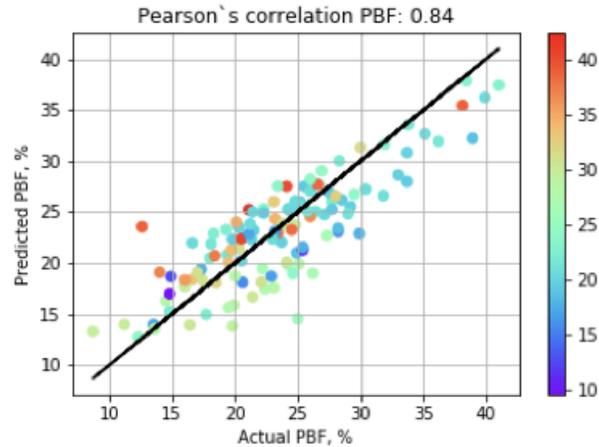

Figure 6: Scatter plot of estimated PBF Vs Actual PBF. Here color represents age distribution.

Figure 5 is the density plot of PBF prediction error (actual-predicted) in the validation data set. The mean absolute accuracy (MAE) and standard deviation (SD) of estimated PBF was 2.73% and 3.34% respectively. From the figure, we see the errors are generally normally distributed with two peaks due to gender differences. However, we observe two outliers at the far left and far right corners of the density plot. This is caused by two distinct cases on our validation data set.

The one on the left (over prediction) is due to the fact that the person was very muscular (body builder) with a very low (12%) PBF. Athletes/bodybuilders tend to have a lower body fat percentage, i.e. around 7% lower for male and 5% lower for females, compared to the non-athletes with similar body height and weight [9]. Less than 1% in our data set was composed of very muscular people. Moreover, we did not do any special treatment for these types of minority dataset. Hence the model performed poorly in that case.

Likewise, in Figure 5, the one on the right is a PBF estimation for a person with a very thin body and small face but with a very high PBF. Hence, the model under-predicted the estimation in this case and got higher prediction error.

Scatter plot in Figure 6 shows the actual PBF (X-axis) vs estimated PBF (Y-axis). The scatter dots are colored by age range (<= 10 years to >= 40 years). Pearson's correlation between predicted and actual PBF was 0.84. From the scatter plots, we can confirm that the prediction accuracy is invariant to the age range of the users.

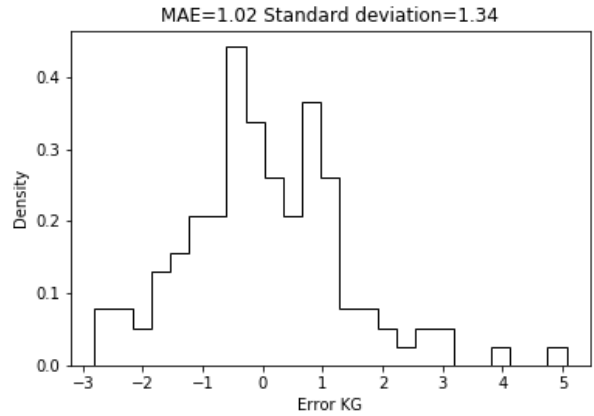

Figure 7: Density plot of SMM estimation error

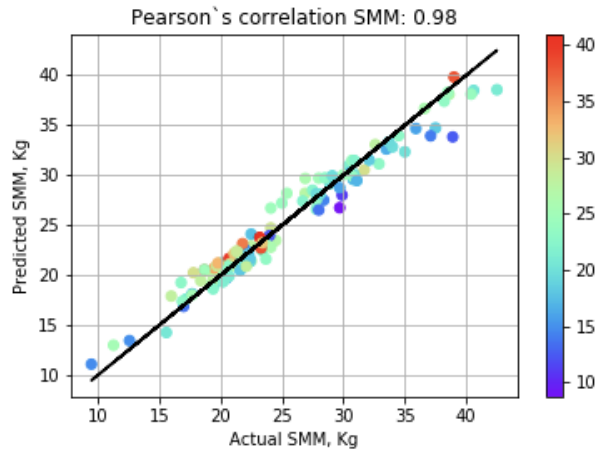

Figure 8: Scatter plot of estimated SMM vs actual SMM. Here color represents age distribution.

Figure 7, is the density plot of SMM prediction error. Regarding SMM prediction accuracy, the MAE and SD of estimated SMM was 1.02kg and standard deviation (SD) of 1.34kg respectively. Since SMM has a very high correlation with body weight (0.86) and height (0.84) (refer fig. 4), the accuracy of predicted SMM is a lot higher compared to the PBF.

Figure 8 is a scatter plot where the dots are colored by age range (<= 10 years to >= 40 years). As in the case of PBF, for SMM too, from the scatter plots we can validate that the age range is not affecting the prediction accuracy. Moreover, in the case of SMM, Pearson's correlation between predicted and actual SMM stands very high at 0.98.

We compared the results of PBF and SMM estimated by our method against the ones that were trained only with height, gender, age and weight as an input, to validate the effectiveness of our proposed technique. For this test, we developed a regression model based on Support Vector Machines, Random Forest, and Gradient Boosting techniques to estimate PBF and SMM as a separate task. The MAE and SD of estimated PBF and SMM from each of those models and our proposed model is tabulated below.

Table 1: Comparison of SMM and PBF estimation using different regression models

| Models | SMM (MAE) | PBF (MAE) | SMM (SD) | PBF (SD) |
| --- | --- | --- | --- | --- |
| Support vector machines | 1.4 kg | 3.6 % | 1.68 kg | 4.32 % |
| Random forest | 1.44 kg | 3.81 % | 1.83 kg | 4.65 % |
| Gradient boosting | 1.18 kg | 3.0 % | 1.47 kg | 3.53 % |
| **Proposed model** | **1.02 kg** | **2.73 %** | **1.34 kg** | **3.34 %** |

From this comparative analysis we can confirm that our proposed method of estimating PBF and SMM by adding clues from facial features is superior at producing better results.

Apart from some extreme cases, our proposed model estimates PBF and SMM with reasonably high accuracy for general users.

## 4. CONCLUSION

In this paper we proposed multimodal multi-task deep neural networks to estimate body composition, namely, body fat percentage and skeletal muscle mass, as a regression task by learning mapping function from facial clues and body measurements (height, gender, age and weight). The mean absolute error of the estimated results of percentage body fat and skeletal muscle mass are 2.7% and 1kg on our validation data set.

We have productionized this algorithm and it is available freely on our body measurement app called 'Bodygram' which can be downloaded from the App Store for iPhone users and from Google Play for Android users.

We are in the process of collecting more variation in the data sets including athletes, non-Japanese people, and people with higher than average weights and heights to further enhance the model to have consistent results for wider demographics and body types.

**References**

[1] Yuki Fukuoka, Takuma Narita, Hiroki Fujita, Tsukasa Morii, Takehiro Sato, Mariko Harada Sassa and Yuichiro Yamada: Importance of physical evaluation using skeletal muscle mass index and body fat percentage to prevent sarcopenia in elderly Japanese diabetes patients. *J Diabetes Investig, 2019*; 10: 322-330

[2] Preethi Srikanthan, Tamara B. Horwich and Chi Hong Sen: Relation of Muscle Mass and Fat Mass to Cardiovascular Disease Mortality. *The American Journal of Cardiology*, Feb 2016

[3] Tanmay Nath, Rexford S. Ahima and Prasanna Santhanam: Body fat predicts exercise capacity in persons with Type 2 Diabetes Mellitus: A machine learning approach. *PLOS ONE*, March 31, 2021

[4] Jian Shou, Pei-Jie Chen and Wei-Hua Xiao: Mechanism of increased risk of insulin resistance in aging skeletal muscle. *Diabetology & Metabolic Syndrome,* volume 12, Article number: 14 (2020)

[5] Steven B. Heymsfield, Abishek Stanley, Angelo piettrobelli and Moonseong Heo: Estimation formulas: What we can learn from them. *Frontiers in Endocrinology*, February, 2020

[6] Christine Mayer, Sonja Windhager, Katrin Schaefer and Philipp Mitteroecker: BMI and WHR are reflected in female facial shape and texture: A geometric morphometric image analysis. *PLOS ONE*, January 2017

[7] Arnab Chanda and Subhodip Chatterjee: Predicting Obesity Using Facial Pictures during COVID-19 Pandemic. *BioMed Research International*, Vol. 2021, Article ID 6696357, 7 pages, 2021

[8] Sebastian Ruder: An Overview of Multi-Task Learning in Deep Neural Networks.
*https://arxiv.org/pdf/1706.05098.pdf*

[9] Damoon Ashtary-Larky, Ali Nazary Vanani, Seyed Ahmad Hosseini, Roya Rafie, Amir Abbasnezhad and Meysam Alipour: Relationship Between the Body Fat Percentage and Anthropometric Measurements in Athletes Compared with Non-Athletes. Zahedan J Res Med Sci, February, 2018

[10] Justine Frija-Masson, Jimmy Mullaert, Emmanuelle Vidal-Petiot, Nathalie Pons-Kerjean, Martin Flamant and Marie-Pia d'Ortho: Accuracy of Smart Scales on Weight and Body Composition: Observational Study. JMIR mHealth and uHealth vol. 9,4 e22487. 30 Apr. 2021